\documentclass[conference]{IEEEtran}

\usepackage{times}
\usepackage[numbers]{natbib}
\usepackage{multicol}
\usepackage[bookmarks=true]{hyperref}

\usepackage{amsmath} 
\usepackage{amssymb}  

\usepackage{graphics} 
\usepackage{epsfig} 
\usepackage{balance}
\newcommand\blfootnote[1]{%
  \begingroup
  \renewcommand\thefootnote{}\footnote{#1}%
  \addtocounter{footnote}{-1}%
  \endgroup
}

\newcommand\Mark[1]{\textsuperscript#1}

\usepackage[framemethod=tikz]{mdframed}

\begin{document}
\title{Ergodic Specifications for Flexible Swarm Control: From User Commands to Persistent Adaptation}
\author{Ahalya Prabhakar, Ian Abraham, Annalisa Taylor, Millicent Schlafly, Katarina Popovic, Giovani Diniz, Brendan Teich, Borislava Simidchieva, Shane Clark, Todd Murphey}

\author{\authorblockN{Ahalya Prabhakar\Mark{1}, Ian Abraham\Mark{1}, Annalisa Taylor\Mark{1}, Millicent Schlafly\Mark{1}, Katarina Popovic\Mark{1}, \\Giovani Diniz\Mark{2}, Brendan Teich\Mark{2} Borislava Simidchieva\Mark{2}, Shane Clark\Mark{2}, Todd Murphey\Mark{1}}
\authorblockA{\Mark{1}Department of Mechanical Engineering\\
Northwestern University,
Evanston, IL 60208}
\authorblockA{\Mark{2}Raytheon BBN Technologies\\
10 Moulton Street Cambridge, MA 02138}
}

 \refstepcounter{figure}
\makeatletter
\let\@oldmaketitle\@maketitle
\renewcommand{\@maketitle}{\@oldmaketitle
  \includegraphics[width=\linewidth]
  {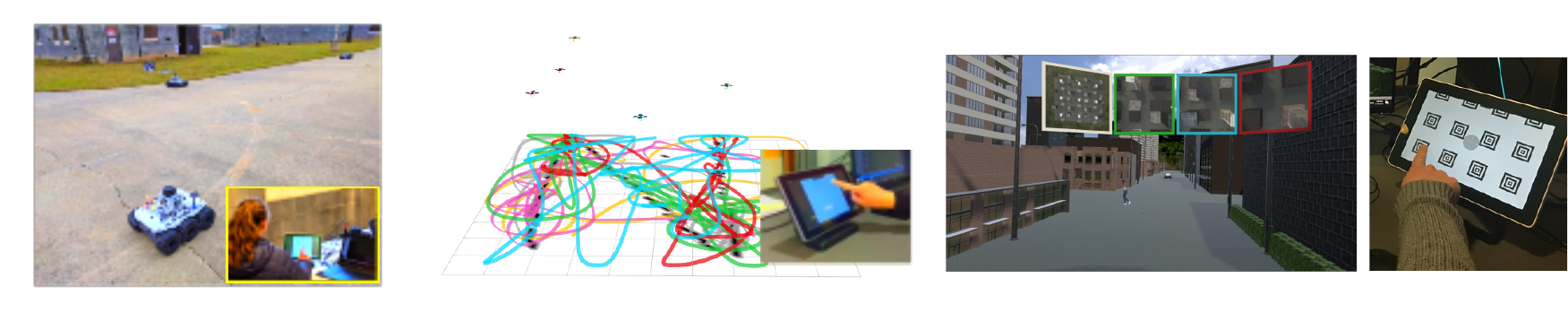} \\[0.35em]
  
 \footnotesize{Fig.~\thefigure. Examples of our proposed method for user-based swarm control from user commands through the TanvasTouch Tablet interface. Here, we show results of dynamic swarm response to user-based inputs in (left) hardware experiments with a swarm of rovers conducted in real-world field testing environment, (middle) simulations with a swarm of quadrotors and (right) simulations of a swarm of drones providing visual feedback to a user in a Unity Virtual Reality game environment developed for end-to-end pipeline validation.
  }
  \label{fig:title} \medskip \vspace{-10pt}}
\makeatother
\maketitle
\thispagestyle{empty}
\pagestyle{empty}

\begin{abstract}

    This paper presents a formulation for swarm control and high-level task planning that is dynamically responsive to user commands and adaptable to environmental changes. We design an end-to-end pipeline from a tactile tablet interface for user commands to onboard control of robotic agents based on decentralized ergodic coverage. Our approach demonstrates reliable and dynamic control of a swarm collective through the use of ergodic specifications for planning and executing agent trajectories as well as responding to user and external inputs. We validate our approach in a virtual reality simulation environment and in real-world experiments at the DARPA OFFSET Urban Swarm Challenge FX3 field tests with a robotic swarm where user-based control of the swarm and mission-based tasks require a dynamic and flexible response to changing conditions and objectives in real-time.\footnote{For multimedia and code, go to \url{https://sites.google.com/view/ergodic-flexible-swarm-control}} \blfootnote{DISTRIBUTION A. Approved for public release: distribution unlimited. Case Number 32600 Work sponsored by DARPA; the views and conclusions contained in this document are those of the authors and not DARPA or the U.S. Government.}

\end{abstract}

\IEEEpeerreviewmaketitle

\section{Introduction}
    One of the biggest problems in multi-agent control of robotic systems is the management and individualized control of the swarm of robots. Specifically, how does one manage a swarm and have the swarm manage itself? For human operators, controlling large numbers of agents increases the cognitive complexity required to manage the swarm~\cite{kolling2015human}, resulting in mental overload~\cite{durantin2014using} or difficulties allocating attention to achieve different tasks~\cite{tessier2012authority,durantin2014using,gateau2016considering}. To enable the human to operate a swarm, some have provided interfaces that abstract the user's control over drone movements, either through hierarchical control~\cite{bevacqua2015mixed} or a swarm-specific programming language~\cite{buzz2016}. Other studies have attempted to make directing a swarm more intuitive by using haptics to provide information about the swarm's current state via a haptic glove~\cite{swarmTouch2019} or 3 DoF Haptic Omega device~\cite{hapticOmega2011}. These methods enable direct mapping between human and swarm motion. However, it is still necessary to develop a method that integrates both a framework to incorporate a user command into the supervision of a swarm and individual robot-level planning algorithms.

    Currently, many swarm planners are inflexible when faced with real-world challenges a swarm of robots would face in the field. Existing methods approach the problem by prespecifying motion behaviors for each robot in swarm formations that collectively accomplish a given task~\cite{giles2017mission,balch1998behavior,setter2015team,kolling2012towards,bevacqua2015mixed}. These methods are rigid to dynamic replanning due to changes in the environment, hardware issues, and number of agents in the swarm. Other planners that attempt to replan based on updates often rely on a ``central control'' to update and assign tasks to each agent in the swarm~\cite{mavrommati2018real}. This approach often makes the system vulnerable to communication issues when agents may not receive updates from the central control~\cite{wei2013agent, sampedro2016flexible}. Our approach attempts to mitigate these issues through a decentralized strategy which is independent of a central control hub and designed around individual agent computation and communication.

    In addition, the method of operator control over the swarm can dramatically affect the joint task performance. In ~\cite{swamy2019scaled}, a supervisor is able to influence one member of a swarm at a time to achieve a task. However, with an increasing number of agents, adjusting swarm behaviour by influencing individual agents becomes less effective. In contrast, defining how the collective swarm behaves using a spatial distribution has been shown to scale favorably. Diaz-Mercado et al. \cite{diazmercado2015density} presents a decentralized, density-based coverage approach which influences multiple robots in a swarm from user commands. In their work, a user defines an area of exploration with a tablet interface. The user-defined density function is broken into partitions for which individual agents are responsible. In contrast, our approach motivates the use of flexible density descriptions where each agent is responsible for coverage of the full area, but can communicate its past and intended trajectory to the other agents. This allows for each agent to prioritize local exploration while ensuring coverage specifications are robust to network dynamics.

    In this work, we consider the following scenarios which address the mentioned problems. We consider a swarm of robots tasked with exploring an environment, while trying to locate and secure easter eggs (EEs) (similar to a scavenger hunt). Once an easter egg is located, available agents should converge on the easter egg to secure it. As more are located, the agents should collectively cover them. At the same time, there are (adversarial) disabling devices (DDs) located around the environment that renders an agent inoperative if they are close to it for a specified short amount of time. The agents have no initial estimate of the number or locations of the easter eggs or disabling devices. In addition, the agents are subject to user command (with the idea being that the human controlling the swarm can assist or divert the swarm). In this scenario, we focus on the problem statement:
    \textbf{How does one generate a swarm control scheme that explores the environment and dynamically adapts to task specifications as information is gained through individual agents and user commands?}

    Thus, in this work our contributions are as follows:
    \begin{itemize}
        \item Develop a formulation to generate swarm control that is dynamically responsive to environment updates and amenable to both user-based and self-managed swarm control
        \item Demonstrate the use of ergodic specifications for defining task allocation as spatial distributions
        \item Demonstrate end-to-end pipeline on a swarm of ground vehicles in a real-world setting at the DARPA OFFSET FX3 field tests
    \end{itemize}

    \begin{figure}
      \centering
      \includegraphics[width=0.35\textwidth]{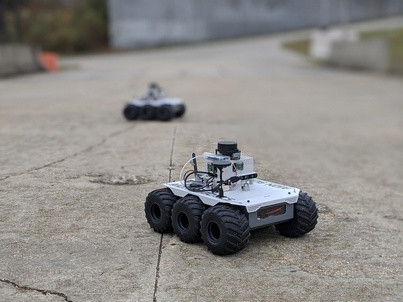}
      \caption{
        The Aion Robotics R1 rover used for the swarm hardware experiments for user-based swarm control and
        mission-based exploration.
        }
      \label{fig:rover_portrait}
    \vspace{-2em}

    \end{figure}

    \begin{figure*}[htb]
      \centering
      \includegraphics[width=.9\textwidth]{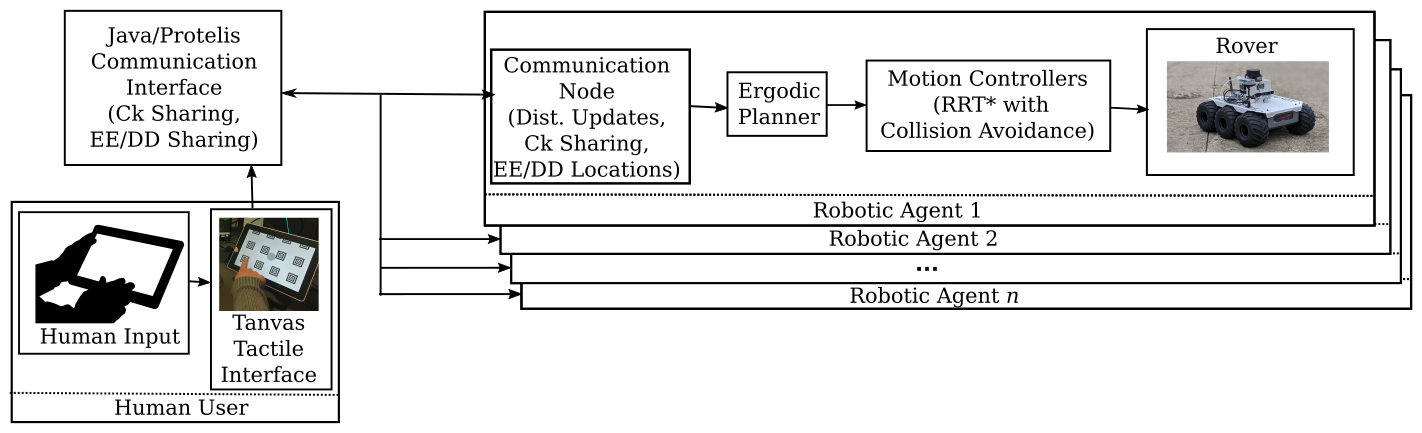}
      \vspace{-1em}
        \caption{
            The overview diagram of the full experimental system architecture. Each robotic agent has its own local ergodic planner and communicates information between the swarm using the Java interface provided by Raytheon BBN. The human user can specify regions of interest through the TanvasTouch tablet interface that communicates the human input through the Java interface using a TCP protocol.
            }
            \label{fig:sys_architecture}
            \vspace{-1em}

      \end{figure*}

      The paper is organized as follows:
      Section \ref{sec:System Overview} describes the full experimental setup, including a description of the interfaces developed and the hardware used in the experiments. Section \ref{sec:Ergodic_Planner} describes the full end-to-end algorithmic formulation, including a brief description of the ergodic planner and the ergodic specifications for user-based control and dynamic environmental response. Section \ref{sec:Results} provides the results from the experiments conducted at DARPA OFFSET FX3 field test as well as some additional simulated examples. Section \ref{sec:Discussion} presents a discussion of these results. Finally, Section \ref{sec:Conclusion} provides some concluding remarks on our system and future work.

\section{System Overview} \label{sec:System Overview}

     \begin{figure}[t!]
      \centering
      \includegraphics[width=0.45\textwidth]{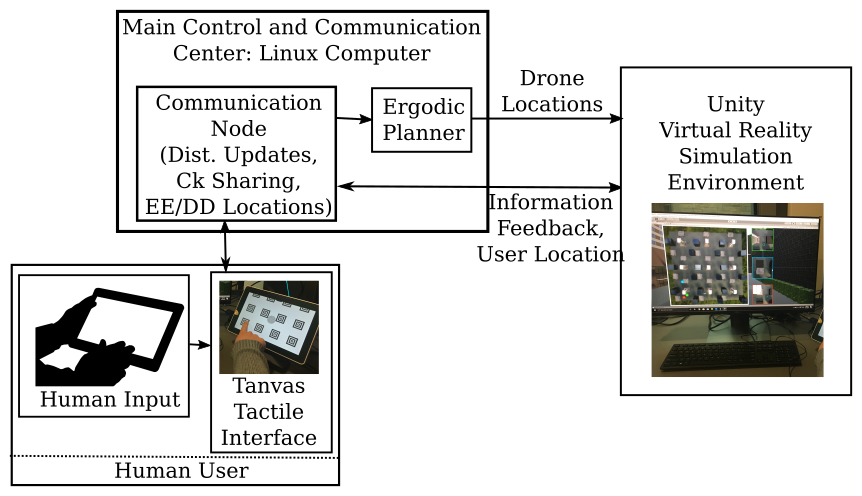}
        \caption{
            The overview diagram of the VR experimental system architecture. The human user can specify regions of
            interest through the TanvasTouch tablet interface that communicates the human input to the main command node
            on the Linux computer using a TCP protocol. The ergodic planner on the Linux computer generates the path for
            the swarm of drones which is communicated to the Unity VR environment over a ROSbridge websocket.
            }
            \label{fig:vrsys_architecture}
    \vspace{-1em}

      \end{figure}

    This section describes the system architecture (shown in Figure \ref{fig:sys_architecture}) for the experiments and
    the virtual reality (VR) simulation setup (shown in Figure \ref{fig:vrsys_architecture}).

    \subsection{Tactile Tablet} \label{sec:Tablet}

        The user sends commands to the rovers and drones using a TanvasTouch monitor (shown in Figure
        \ref{fig:vr_system}), which is designed to generate surface haptics. The TanvasTouch renders textures on the
        smooth screen by modulating the friction underneath the user's fingertip. We utilize the haptic features of the
        TanvasTouch to render a haptic aerial map of the the environment, representing different objects such as
        buildings with different textures. This enables the user to orient themselves in their environment and send
        commands without needing to look down.

        Using the TanvasTouch, the user can specify regions of exploratory interest by simply double-tapping the screen,
        shading the region on the TanvasTouch, and double-tapping the screen again. At the second double-tap, the
        coordinates are transformed, represented as a distribution, and sent over a TCP socket to each agent listening
        to the commands. Auditory feedback is provided for the double-taps. When the TanvasTouch is used with the
        virtual reality interface, the haptic display dynamically updates according to the person's position and
        orientation.

    \subsection{Unity VR Interface}

\begin{figure}
          \centering
          \includegraphics[width=0.4\textwidth]{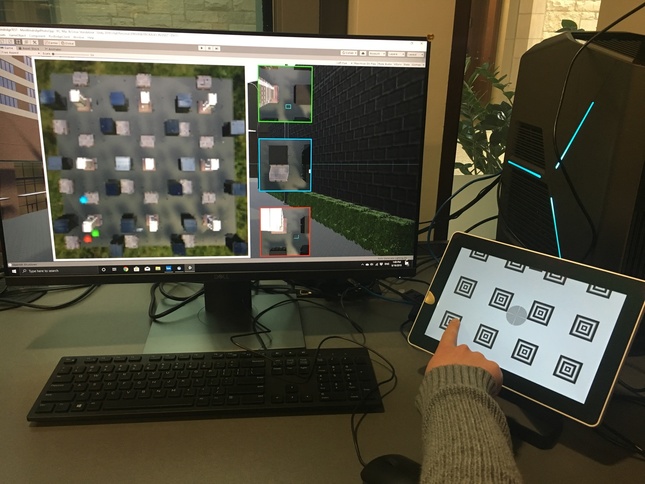}
          \caption{
            The Virtual Reality testbed developed to validate the full pipeline of swarm control, from a user's
                input through the Tanvas tablet interface to ergodic swarm control in a Unity VR Environment. The VR environment displays the visual feedback from each drone in the swarm to a user moving through the environment, along with an aerial view of the drone's locations and any objects of interest they locate.
                }
                \label{fig:vr_system}
          \vspace{-2em}

        \end{figure}

        For the purpose of validating our system ahead of the field test, we developed an experimental urban environment
        testbed using the Unity game engine (shown in Figure \ref{fig:vr_system}). This virtual reality (VR) environment
        was used to validate the full system architecture of the ergodic formulation with multiple two-way communication
        channels (shown in Figure \ref{fig:vrsys_architecture}) prior to implementation on hardware.

        The ergodic algorithm ran on a Linux machine running Ubuntu 18.04 with an Intel 8 core i7-8550U CPU 1.80GHz
        processor using ROS (Robot Operating System, version melodic). It communicated the resulting swarm controls to
        the Unity system on Windows using a ROSbridge websocket interface. The ergodic algorithm controlling the swarm
        was initially driven for uniform exploration. As exploration progressed, the drones communicated important
        information (i.e., locations of easter eggs or antagonistic devices) identified from the drone's visual feedback
        to dynamically update the swarm behavior. For the purpose of testing, this feedback assumed perfect semantic
        understanding of the identified items in the field. The operator used the HTC Vive controller to move through the VR environment. In addition, the user used the TanvasTouch to generate
        desired area of exploration.  The operator location from Unity was sent to the TanvasTouch to render the
        appropriate haptic feedback on the display. In return, the operator’s input from the tablet, translated to the
        array of spatial points, was sent to each agent running an individual instance of the algorithm to dynamically update the behavior.

    \subsection{Experimental Setup}

        This section describes the full experimental system architecture shown in Figure \ref{fig:sys_architecture}. The
        swarm hardware used in the field tests consisted of 4 Aion Robotics R1 rovers (shown in Figure \ref{fig:rover_portrait}. The rover includes an NVIDIA
        Jetson TX-2 embedded computing device and Pixhawk 2.1 controller with a HERE GPS unit running ArduRover 3.2.
        Onboard sensors include a RPLidar A1M8 2D 360 Lidar and an Intel RealSense D435i depth camera. Motion planning
        on the rover used the decentralized ergodic planner described below in Section \ref{sec:Ergodic_Planner} wrapped
        around an RT-RRT*~\cite{naderi2015rt} for real time path planning and obstacle avoidance developed by Raytheon BBN Technologies. The
        ergodic planner algorithm ran locally on the NVIDIA TX-2 of each individual rover running a local ROS network. The
        rovers in the swarm communicated with each other and received swarm-level commands over a local LTE network
        through a Java interface developed by the Raytheon BBN team.

        The experiments were conducted at the DARPA OFFSET field tests on a smaller range area containing a combination of grassy terrain and concrete
        sidewalks. The bounding box within which the rovers operated was defined by GPS coordinates taken from the
        Java interface provided by Raytheon BBN. User commands were communicated from the TanvasTouch tablet
        interface connected through the Java interface to the robotic agents using a TCP protocol.

        The following section derives the ergodic planner and specifications for control of a swarm of heterogenous robots.

\section{Algorithm Overview} \label{sec:Ergodic_Planner}

In this section, we introduce and describe ergodicity as a concept for converting spatially distributed task information into temporally driving robot motion. In addition, we present the decentralized ergodic planning algorithm for a networked set of heterogenous agents, adapted from the decentralized ergodic controller in \cite{abraham2018decentralized}.

    \subsection{Ergodicity and the Ergodic Metric}

        In this work, ergodicity is used to compare the temporal statistics of a swarm of robotic agents to a desired spatial distribution. In order to define ergodicity and the ergodic specification that we use, let us assume that the state of a single robotic agent at time $t$ is given by $x(t) : \mathbb{R}^+ \to \mathbb{R}^n$ and the controls to the robot at time $t$ are defined as $u(t) : \mathbb{R}^+ \to \mathbb{R}^m$. The dynamics of the robot are then defined to be the control-affine dynamical system of the form

        \vspace{-1em}
        {\small
        \begin{equation} \label{eq:robot_dynamics}
            \dot{x}(t) = f(x(t),u(t)) = g(x(t)) + h(x(t)) u(t)
        \end{equation}}
      \vspace{-1em}
      
        where $g(x) : \mathbb{R}^n \to \mathbb{R}^n$ is the free, unactuated dynamics of the robot, and $h(x):
        \mathbb{R}^n \to \mathbb{R}^{n \times m}$ is the dynamic control response subject to input $u(t)$. Let us
        now define the robot's time-averaged statistics $c(s, x(t))$ for a trajectory $x(t)$ (i.e., the statistics describing
        where the robot spends most of its time) for some time interval $t \in \left[ t_i, t_i + T\right]$ as

        {\small
        \begin{equation}\label{eq:time_avg_stats}
            c(s, x(t)) = \frac{1}{T}\int_{t_i}^{t_i+T} \delta (s - x_v(t)) dt,
          \end{equation}}

        where $\delta$ is a Dirac delta function, $T \in \mathbb{R}^+$ is the time horizon, $t_i \in \mathbb{R}^+$ is
        the $i^\text{th}$ sampling time, $s \in S^v$ is a point in the exploration space $S^v$, and $x_v(t) \in S^v \cap \mathbb{R}^n$ is the state that intersects with the exploration 
        space, where $v \le n$. An ergodic metric~\cite{mathew2011metrics} which relates the time-averaged distributions
        $c(s,x(t))$ and arbitrary spatial distribution $\phi(s)$ is:

                \vspace{-1em}
        {\small
        \begin{align} \label{eq:ergodic_metric}
            \mathcal{E}(x(t)) & = q \,\sum_{k \in \mathbb{N}^v} \Lambda_k \left(c_k -\phi_k \right)^2   \\
            & = q \, \sum_{k \in \mathbb{N}^v} \left( \frac{1}{T} \int_{t_i}^{t_i + T} F_k(x(t)) dt - \phi_k \right)^2 \nonumber
        \end{align}}
              \vspace{-1em}

      where
      
        \vspace{-1em}
      {\small
        \begin{equation*}
            \phi_k =  \int_{\mathcal{X}_v} \phi(s) F_k(s) ds,
          \end{equation*}}
                \vspace{-1em}

        $q \in \mathbb{R}^+$ is a scalar weight on the metric, and $c_k, \phi_k$ are the Fourier decompositions \footnote{The
        cosine basis function is used, however, any choice of basis function $F_k$ can be used.} of $c(s,x(t))$ and
      $\phi(s)$ with

        {\small
        \begin{equation*}
            F_k(x) = \frac{1}{h_k}\prod_{i=1}^v \cos \left( \frac{k_i \pi x_i}{L_i} \right)
          \end{equation*}}

          being the cosine basis function for a given coefficient $k \in \mathbb{N}^v$, $h_k$ is a normalization factor
        defined in~\cite{mathew2011metrics}, and $\Lambda_k = (1 + \Vert k \Vert^2)^{-\frac{v+1}{2}}$ are weights on the
        frequency coefficients. A robot whose trajectory $x(t)$ minimizes (\ref{eq:ergodic_metric}) as $t\to \infty$ is
        then said to be optimally ergodic with respect to the target distribution. That is, the robot spends time in
        regions of the exploration space proportional to the spatial statistical measure in the exploration space.

        By directly generating trajectories that minimize the ergodic metric, what we obtain is a method for specifying how
        long a single agent should spend time in particular regions of task space. This is important for dynamically
        changing tasks and introducing user commands as the task definition can be composed into a multimodal
        distribution. Minimizing the ergodic metric thus avoids issues often faced with multimodal optimizations as
        the robot will allocate proportional amounts of its time within some allotted time depending on the measure of
        importance specified by all the elements that are desirable (e.g., easter eggs and overriding user commands).

        In the following section we show that the benefit of the ergodic metric is not only with the flexibility of
        specifying tasks but also with allowing for dynamic decentralized network of agents.

        \begin{figure*}[htb]
          \centering
          \includegraphics[width=0.95\textwidth]{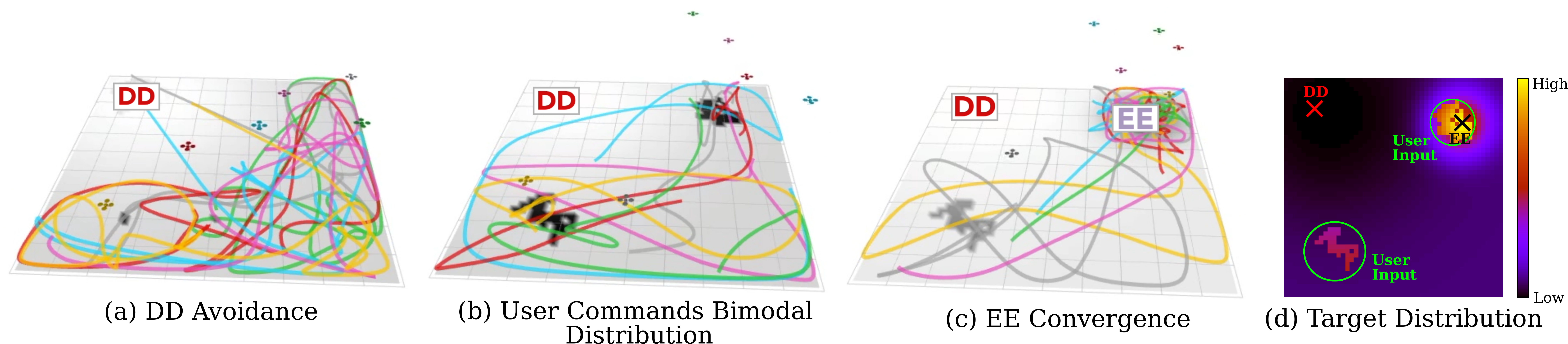}
          \vspace{-.75em}
            \caption{
            Simulation of a swarm dynamically adapting to the environment while also responding to user commands. Shown are the robots' trajectories in response to the different stimuli. In (a), when the swarm discovers a DD, the agents cover the rest of the workspace while avoiding that location. When a user inputs a bimodal distribution for the swarm to cover (shown as the dark region on the map), the swarm responds to the user commands, while continuing to avoid the DD location (in (b)). (c) shows the response to the swarm locating an EE in the environment. The swarm simultaneously converges on the EE to secure it, covers the user inputs, and avoids the DD location. (d) shows the resulting target distribution for the combined tasks. The x labels mark the DD location (in red) and EE location (in black) and the circles in green highlight the user input from the Tanvas interface.                 %
            }
            \label{fig:combsim}
            \vspace{-1.5em}

          \end{figure*}

    \subsection{Decentralized Ergodic Control} \label{subsec:decentralized-ergodic-control-using-consensus}

        The utility of the ergodic metric comes not from the flexible specification, but in how the contribution
        of an agent's motion in solving the metric is calculated. In this section, we provide an overview of decentralized ergodic control \cite{abraham2018decentralized}.

        Consider a set of $N$ agents with state $x(t) = \left[ x_1(t)^\top, x_2(t)^\top, \ldots, x_N(t)^\top\right]^\top :
        \mathbb{R}^+ \to \mathbb{R}^{n N}$. For readability we consider a homogeneous set of agents with the
        same state dimension $x_i(t) \in \mathbb{R}^n$, but, this analysis can be done for a heterogeneous set of
        agents. Since the collective temporal statistics of the swarm are calculated with only the shared $c_k$ values, for the heterogeneous case, the ergodic control (\ref{eq:policy_independence}) is calculated for each agent independently with respect to their individual dynamics. The multi-agent system's contribution to the time-averaged statistics $c_k$ can be rewritten as

        \vspace{-1em}
        {\small
        \begin{align}\label{eq:centralized_ck}
            c_k & = \frac{1}{N} \sum_{j=1}^N
                    \frac{1}{T} \int_{t_i}^{t_i + T} F_k(x_j(t)) dt \nonumber \\
            & = \frac{1}{T} \int_{t_i}^{t_i+T} \tilde{F}_k(x(t)) dt
        \end{align}}
        \vspace{-1em}

        where $\tilde{F}_k(x(t)) = \frac{1}{N}\sum_j F_k(x_j(t))$.
        We can additionally show that each agent can generate an independent action that contributes to minimizing the
        global ergodic metric. Let us first define the dynamics of the collective multi-agent system as

                \vspace{-1em}
        {\small
        \begin{align} \label{eq:collective_dynamics}
            \dot{x} & = f(x,u) = g(x) + h(x) u \nonumber\\
            & = \begin{bmatrix}
            g_1(x_1) \\
            g_2(x_2) \\
            \vdots \\
            g_N(x_N)
            \end{bmatrix} +
             \begin{bmatrix}
            h_1(x_1) & \ldots & 0\\
            \vdots& \ddots & \\
            0 & & h_N(x_N)
            \end{bmatrix} u.
        \end{align}}

      We can calculate the adjoint variable of the ergodic objective function as

        {\small
        \begin{equation}\label{eq:decentralized_adjoint}
            \dot{\rho} = -2 \frac{q}{T}\sum_{k \in \mathbb{N}^v} \Lambda_k \left( c_k - \phi_k \right) \frac{\partial \tilde{F}_k}{\partial x} - \frac{\partial f}{\partial x}^\top \rho
          \end{equation}}
          
        where

        {\small
        \begin{equation*}
            \frac{\partial \tilde{F}_k}{\partial x} = \frac{1}{N} \begin{bmatrix}
            \frac{\partial F_k (x_1)}{\partial x_1} \\
            \vdots \\
            \frac{\partial F_k(x_N)}{\partial x_N}
            \end{bmatrix}
            \text{ and }
            \frac{\partial f}{\partial x} =
            \begin{bmatrix}
            \frac{\partial f_1}{\partial x_1} & 0 & \ldots & 0 \\
            0 & \frac{\partial f_{2}}{\partial x_{2}} \\
            \vdots & & \ddots  & \\
            0 &  & & \frac{\partial f_N}{\partial x_N}
            \end{bmatrix}
          \end{equation*}}

        is block diagonal. Because each agent's dynamics are independent of each other, (\ref{eq:decentralized_adjoint})
        can be written independently for each agent as

        {\small
        \begin{equation*}
            \dot{\rho}_j = -2\frac{q}{T N} \sum_{k \in \mathbb{N}^v} \Lambda_k (c_k - \phi_k) \frac{\partial F_k(x_j)}{\partial x_j} - \frac{\partial f_j}{\partial x_j}^\top \rho_j.
          \end{equation*}}

        As a result, (following the work in~\cite{mavrommati2018real}) we can define a controller for the collective swarm that
        minimizes the ergodic metric:

        {\small
        \begin{multline} \label{eq:expanded_control}
            \begin{bmatrix}
            u_{\star,1} (t) \\
            \vdots \\
            u_{\star,_N}(t) \\
            \end{bmatrix}
            =
            -R^{-1}
             \begin{bmatrix}
            h_1(x_1) & \ldots & 0\\
            \vdots& \ddots & \\
            0 & & h_N(x_N)
            \end{bmatrix} ^\top
            \begin{bmatrix}
            \rho_1 (t) \\
            \vdots \\
            \rho_N(t)
            \end{bmatrix}
          \end{multline}}

        where $R\in\mathbb{R}^{mN \times mN}$ is a positive definite weight matrix and $mN$ is the size of the swarm
        system control input. Since $h(x)$ is block diagonal,  (\ref{eq:expanded_control}) becomes

        {\small
        \begin{equation}\label{eq:policy_independence}
            u_{\star,j} (t) = -R_j^{-1} h_j(x_j)^T \rho_j(t)
          \end{equation}}
        \vspace{-1em}

        for each agent $j \in \left[ 1, \ldots, N \right]$ and $R_j \in \mathbb{R}^{m \times m}$. Note that the
        $j^\text{th}$ agent does not depend on the $i^\text{th}$ agent. All that is required is that we communicate the
        $c_k$ values between each agent before computing the controller in order to obtain the time-averaged statistics
        of each agent. Each agent then has their own specification of the task target distribution $\phi(x)$ (that is
        parameterized and communicated between agents as well, depending on the underlying type of agent). Since the computational burden lies in computing the ergodic control, which is fully decentralized for each individual agent, the computational complexity of the ergodic controller only depends on the number of decompositions which  scales with the dimensionality of the search environment with order $\mathcal{O}(v^k)$, but does not depend on number of agents or topology. 

        Under fully connected, healthy, network connection, we can assume typical consensus properties. However, under the worse
        case network (e.g., a single agent is left detached from the network) the single agent is fully responsible for minimizing
        their own ergodic metric (and ideally spending time in regions of the task space proportional to their importance). The
        added benefit of communication helps the individual agent minimize the energy expended by moving across the task space.

        In the following section, we show how we adapt the decentralized ergodic control with local planners for obstacle avoidance,
        and low-level control.

                \begin{figure*}[htb]
                  \centering
                  \includegraphics[width=0.75\textwidth]{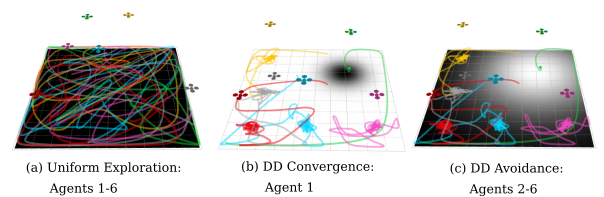}
                  \vspace{-1em}
                    \caption{
                    Swarm dynamically responding to locating a disabling device (DD) and responding to it based
                    on their heterogeneous capabilities. Presented is the robots' trajectories of the DD
                    blocker agent (Agent 1 shown in green) and the rest of the swarm (Agents 2-6 other than green) as they dynamically
                    respond to environmental stimuli. In (a), the swarm uniformly covers the
                    workspace regardless of their individual capabilities. When a DD location is registered, the swarm updates its
                    distribution based on their assigned tasks. Target distribution of (b) DD blocker agent with it converging on the DD and (c) all other agents avoiding the DD.
                    }
                    \label{fig:hetsim}
                    \vspace{-2em}
                  \end{figure*}

        \subsection{Interfacing with RT-RRT$^\star$}
            The control signal $u(t)$ for each agent is converted to kinematic input $\dot{x}(t)$ where the forward
            simulation $x(t)$ for each individual agent is supplied to the RT-RRT$^\star$ low-level planner as a target
            path. This allows for the high-level ergodic planner to avoid needing to consider obstacle avoidance and
            divides the computational load into two segments: low-level planning for robot control and obstacle
            avoidance, and high-level task adaptation with the ergodic planner which involves network communication,
            information sharing, and adapting to user commands. As the ergodic planner is temporally driven (i.e., the
            amount of time spent in a region directly impacts the following planned behavior),  any regions that can
            not be visited due to obstacles will eventually be planned around as the ergodic specification will want to
            explore persistently depending on the spatial measure defined by the target distribution (and the task
            specification).

    \subsection{High-level Planning for Dynamic-Task Adaptation} \label{sec:Task Specs}

        In this section, we define how the tasks are represented as spatial measures defined by $\phi(x)$. Furthermore,
        we specify how user commands are combined with existing task specifications to enable multimodal descriptions
        of where agents need to be allocated.

        \subsubsection{Dynamic Environment Response}
            We focus on two main scenarios: reallocating priority to a given region when an easter egg (EE) is
            discovered and generating a region of avoidance if a disabling device (DD) is discovered. In both of these
            cases, the location of the EE or DD, when discovered, is communicated to each agent in the network. The
            location of the environment elements is introduced into our specification of $\phi(x)$ by parameterizing the
            distribution as a multimodel sum of Gaussians:

            {\small
            \begin{align}
                \phi(x) &= \frac{1}{\eta_a} \sum_{i} a_i \exp \left( - \frac{1}{2}\Vert x - x_\text{EE}\Vert^2_{\Sigma^{-1}}\right) \nonumber \\
                        &\ +   \frac{1}{\eta_b}\sum_{j} b_j \left(1-\exp \left( - \frac{1}{2}\Vert x - x_\text{DD}\Vert^2_{\Sigma^{-1}}\right) \right)
            \end{align}}
            where $\eta_{a,b}$ are normalization factors,  $\sum_i a_i = 1$ and $\sum_j b_j = 1$, and $x_\text{EE}, x_\text{DD}$ are the locations of the EE and
            DD respectively. The parameter $\Sigma$ is the width of the region of attraction (or repulsion) that can be
            tuned based on the size of the task space and the desired granularity. We used $\Sigma = \text{diag}(0.01,
            0.01)$ for both EE and DD. This representation generates high importance regions where there is an EE and
            low regions (avoidance regions) where there is a DD. The resulting distribution is then normalized and
            represented using $10$ Fourier coefficients in each exploratory dimension. Workspace coordinates are
            transformed and scaled to a bounding box of size $[0,1]^2$ for numerical stability.

        \subsubsection{User Command with Tablet Interface}

            The tablet interface described in Section \ref{sec:Tablet} transmits a set of desired points on the
            workspace for the swarm to prioritize. The spatial distribution is generated by assigning the highest
            priority value of 1 at each of those points in a discretized workspace (since the decomposition of $\phi(x)$
            is done numerically, the calculations do not change) and random noise between $[0,.001)$ at every other point
            to generate minimal coverage over the rest of the workspace. The user inputs are added to the parametrized
            distribution for combined use with the EE and DD task. As before, the resulting distribution is then normalized
            and represented using 10 Fourier coefficients in each exploratory dimension and the workspace coordinates are
            transformed and scaled to a bounding box of size $[0,1]^2$.

\section{Results} \label{sec:Results}

    We first demonstrate a simulated example of the tasks mentioned in Section \ref{sec:Task Specs} with a swarm of
    quadcopters. We then demonstrate our formulation on a swarm of ground vehicles for the different task scenarios
    described in Section \ref{sec:Task Specs}. Because of hardware issues between the different trials, the different
    scenarios involve different numbers of robots, based on what was feasible at that moment. The algorithm itself is
    not affected by the number of agents in the swarm (which we show in simulation as well).

    \subsection{Simulated}

        First, we simulate a swarm of 6 agents exploring and dynamically adapting their control response simultaneously to environmental stimuli and user commands through the Tanvas tablet interface. Figure \ref{fig:combsim}a shows the swarm locating and avoiding a DD while exploring the workspace. The user then sends a bimodal input to the swarm. In Figure \ref{fig:combsim}b, the swarm simultaneously responds to the user's bimodal coverage command, while continuing the avoidance of the DD location. Figure \ref{fig:combsim}c shows the swarm responding to a discovered EE. The swarm simultaneously converges on the EE location to secure it, but continues to cover the other peak to satisfy the user commands, all while avoiding the DD location (the resulting target coverage distribution is shown in Fig. \ref{fig:combsim}d).

        Next, we simulate a swarm of heterogeneous agents exploring and dynamically adapting their control response to
        environmental stimuli based on their individual capabilities. We simulate a swarm of 6 agents--- 5 regular agents tasked
        to explore the search-space and 1 agent with a DD blocking capability that renders a DD ineffective.  We simulate an
        agent discovering a DD location and communicating it to the swarm. The regular agents avoid the DD location, while the
        DD blocker agent instead converges to the location to make it safe for any agent that may come close to it.

        In Figure \ref{fig:hetsim}a, the swarm uniformly covers the workspace regardless of the
        capabilities. When a DD location is discovered, the DD blocker agent converges on
        the DD location to render it safe, shown in Figure \ref{fig:hetsim}b, while the rest of
        the swarm continues uniformly exploring the rest of the workspace while avoiding the main DD location (in Figure
        \ref{fig:hetsim}c).

    \subsection{User Interface Results}
        \begin{figure}
          \centering
          \includegraphics[width=0.5\textwidth]{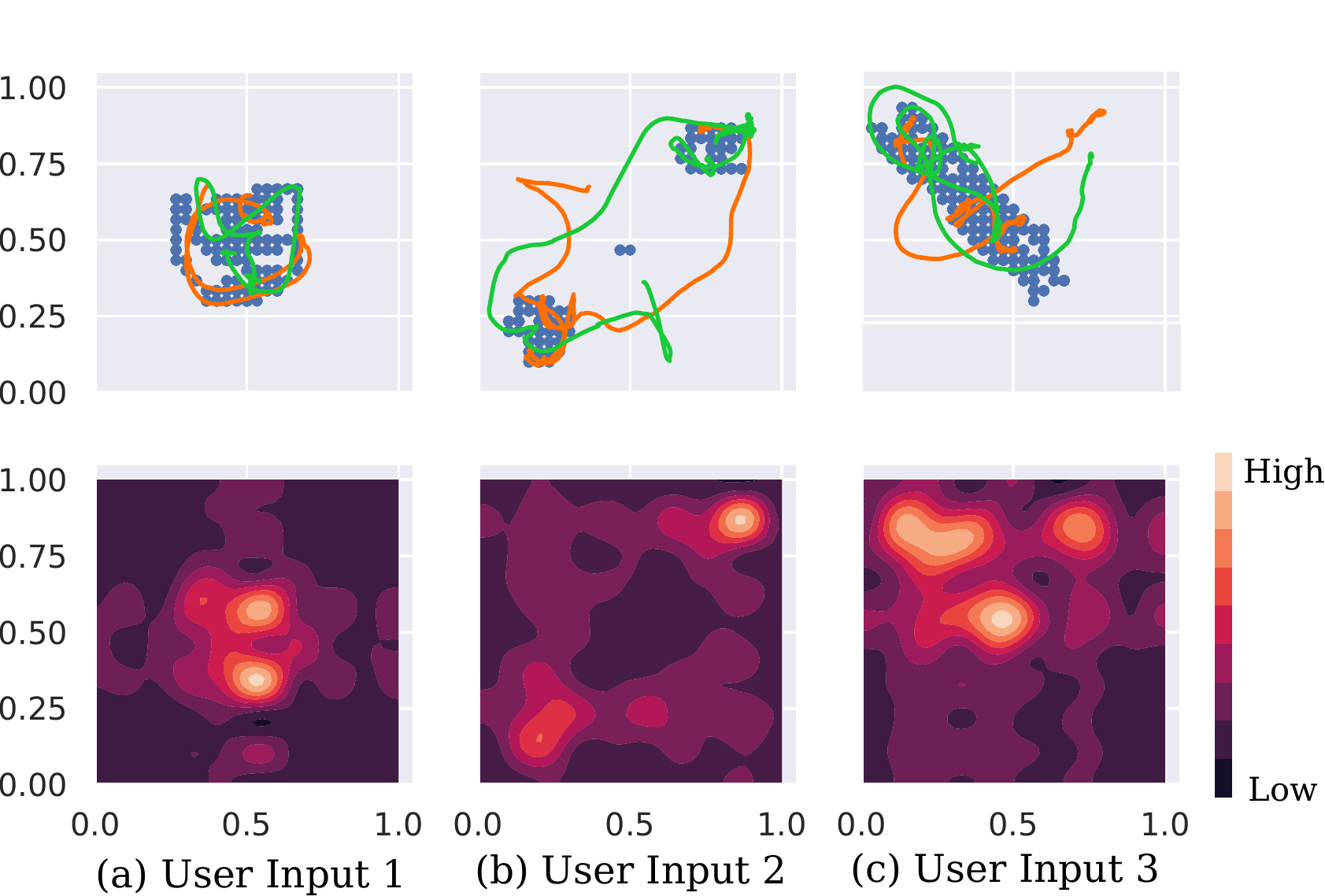}
          \vspace{-2em}
            \caption{Swarm dynamically responding to a user command as communicated through the TanvasTouch tablet interface. The figures show the (top) time-lapse robots' trajectories (shown in orange and green) and (bottom) corresponding Fourier reconstruction of the collective time-averaged trajectories each time the user's input distribution is updated (a,b,c), represented by blue dots highlighting regions of interest in the top figure. Each time the user input is updated with a new distribution, the robots prioritize collectively covering the new target distribution.
            }
            \label{fig:tanvas_timelapse}
            \vspace{-2em}
        \end{figure}

        \begin{figure*}[tb]
          \centering
          \includegraphics[width=0.9\textwidth]{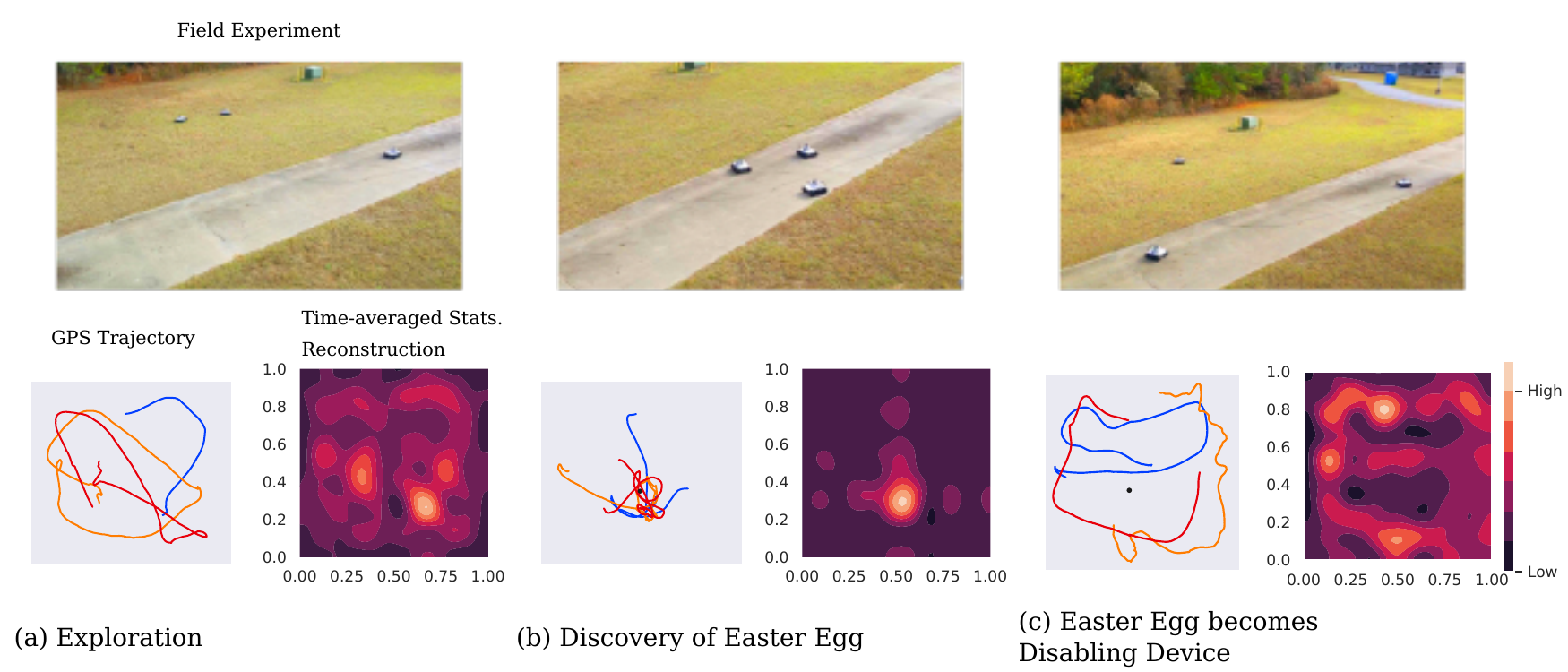}
                    \vspace{-1em}

            \caption{
                Swarm exploring for a discovering of an easter egg (EE) (a) (Top) illustrates the field experiment
                results with a set of 3 agents. (Bottom left) is the rovers' trajectories (in red, orange, and blue) and
                (bottom right) the corresponding Fourier reconstruction of the collective time-averaged trajectories as
                they dynamically respond to environmental stimuli. (b) The same swarm is now dynamically responding to locating (EE)
                location to prioritize. The EE is converted into a disabling device (DD) creating an avoidance
                region around that point (c).
            }
            \label{ied_timelapse}
            \vspace{-1.25em}
        \end{figure*}

        We first demonstrate the user command portion of our end-to-end pipleline from the tablet interface we
        developed to communicate a user's commands to the dynamic update of the swarm allocation using the ergodic
        specification. In Fig.~\ref{fig:tanvas_timelapse} we present a time-lapse of a sequence of user commands to the
        set of rovers. Through the ergodic specification and communication, each agent is able to allocate the time
        spent in each of the target regions specified by the user commands (as shown in the bottom of
        Fig.~\ref{fig:tanvas_timelapse} as the time-averaged statistic reconstruction).

        The rovers successfully adapted to the operator's dynamically changing coverage priorities. As they received new
        inputs through the TanvasTouch tablet interface, they updated their target coverage distributions and explored
        proportional to their importance. Furthermore, though the number of agents in the swarm reduced by 2 due to
        hardware issues, the ergodic algorithm naturally accommodated these changes, dynamically updating the generated
        motion paths of the remaining agents in the swarm so that their collective trajectories still satisfied the
        coverage goals.

    \subsection{Environmental Responses}

        Next, we demonstrate the swarm exploring and dynamically adapting their control response to environmental
        stimuli with 4 ground vehicles. We simulate a rover discovering an EE location and communicating the
        location to the swarm (enabling each agent to adapt their own task specification). We then simulate the
        swarm identifying in that same location a DD that needs to be avoided.

        Figure \ref{ied_timelapse} shows the swarm results at each stage of the scenario described above. The rovers
        start out uniformly exploring the task space (Fig.~\ref{ied_timelapse}(a)) until an easter egg (EE) is
        discovered (Fig~\ref{ied_timelapse}(b)). It is at this moment that the agent that discovered the EE shared
        the location to the other agents, allowing the swarm to adapt their own tasks (locally) and assist the swarm
        to maintain a perimeter around the EE. Note that when covering the location, the rovers do not move to the
        peak and wait--- instead, they constantly generate persistent, exploratory actions over the region (see
        bottom time-averaged statistics reconstruction Fig~\ref{ied_timelapse}). We then simulate the EE becoming a
        DD (so that the agents have to avoid the target region) (see Fig.~\ref{ied_timelapse}(c)). As they received
        the information over the network, they dynamically update their target distributions to explore everywhere
        else around the workspace. Each agent diverged from that location and explored uniformly around the rest of
        the task space (allocating different paths as they communicate their respective time-averaged statistics ($c_k$)).

  \section{Discussion}
        \label{sec:Discussion}

        In the previous section, we present results that validate our formulation for generating swarm control for
        mission-based scenarios, both in simulation and in real-world experimentation. Here, we further discuss these
        results, as well as observations and takeaways from the experiments performed at the DARPA OFFSET FX3 field
        tests.

        \subsection{Persistent Exploration as Applied to Human Cognitive Load}

        \begin{figure*}
          \label{fig:vr_pipeline}
          \centering
          \includegraphics[width=.7\textwidth]{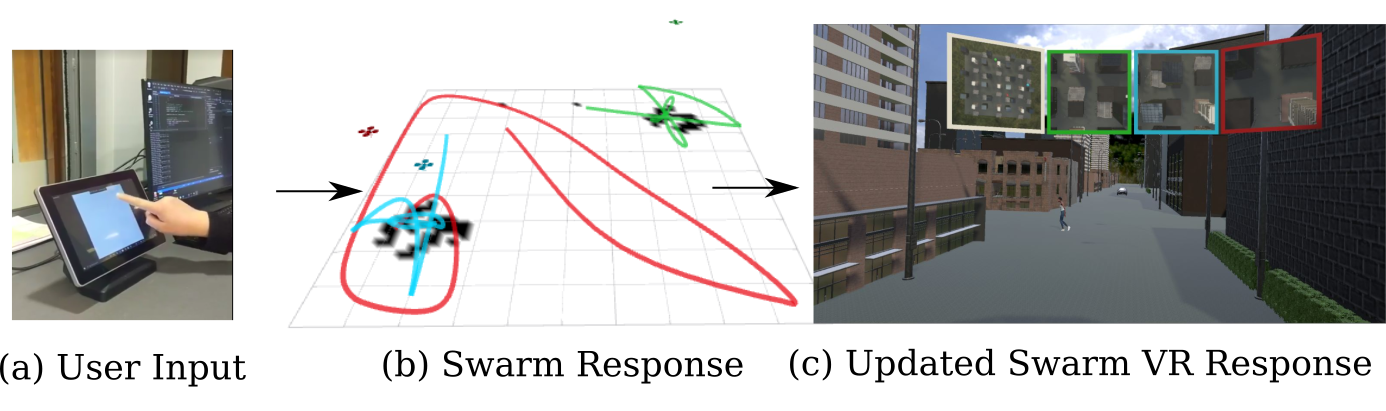}
          \vspace{-1em}
          \caption{Results from the end-to-end VR system pipeline. (Left) The user sends a command to the swarm using the TanvasTouch tablet interface, (Middle) the dynamic response of the swarm to the user input in simulation and (Right) the drone exploration is updated in the  the Unity VR environment from the ergodic planner. The VR environment provides visual feedback from each drone to the user.
          }
          \vspace{-1.75em}
        \end{figure*}

            The decentralized ergodic specification enables each agent to constantly generate actions that minimize the difference between the time-averaged trajectories of the swarm and the target coverage distribution. This type of formulation allows the swarm to manage itself (through communication) and performs as best as a single agent would in the worse-case scenario. A human operator would then not require the cognitive capacity to manage each individual agent constantly to ensure successful task completion.

The virtual reality system (Fig~\ref{fig:vr_pipeline}) presents a testbed for the full end-to-end  pipeline for our proposed system for managing swarms. The user is able to move through the VR environment, while using the tactile tablet to specify regions of interest for the swarm to prioritize exploration when desired, without needing to constantly supervise the swarm behavior. Our goal is to use this system to test human cognitive load in the VR environment where we will test the effects of multitasking on a user attempting to simultaneously manage a swarm in the environment while accomplishing a task.
\subsection{Algorithmic Robustness}
The formulation was particularly robust to many of the real-world issues we faced at the field tests. In particular, the decentralized ergodic specification of the optimization problem allowed the individual agents to generate solutions that adapted to the number of agents in communication with it. This characteristic led to a high degree of robustness to many of the hardware and network issues we faced during experimentation. As the number of rovers in the swarm decreased from rover hardware failure, the ergodic algorithm naturally accommodated these changes, dynamically updating the generated motion paths of the remaining agents in the swarm to collectively satisfy the allocation goals.
This flexibility to dynamically changing numbers of agents also led to compensation for network communication issues. As the number of agents in the collective swarm changed due to network dropout, the ergodic algorithm similarly adapted, spreading the task coverage over the remaining agents and readjusting as agents recovered network communication. While this does result in multiple agents doing the same task, this behavior is often desirable and viewed as task robustness: each agent would assume that the desired task (specified by the target distribution) is still a priority and its sole job was to finish the task. Furthermore, the constant optimization towards the target distribution naturally addressed many of the real-world swarm issues we faced in the field tests due to deadlock between rovers, inaccessible regions or slippage due to the environmental terrain.
\section{Conclusion}
\label{sec:Conclusion}

    This work presents an end-to-end pipeline for managing swarm control for high-level tasks and adaptation to external user commands. We develop a tactile tablet interface for translating user priorities to the swarm. We develop a decentralized ergodic formulation for swarm control for high-level tasks that can dynamically adapt to the demands of a user controlling the swarm and external task priorities from environmental information. We successfully demonstrate our pipeline both in a VR simulated environment and in physical hardware experiments, illustrating the formulation's robustness to many challenges faced in real-world scenarios and its flexibility to dynamically changing needs and priorities for real-time control of a robot swarm.

The current implementation still has its limitations. While our experimental system currently uses a homogeneous swarm of robotic agents, future work will extend our simulation results to real-world experiments, generating different dynamic responses to environmental updates for heterogeneous agents with different capabilities. Additionally, our prototype implementation separated the different high-level tasks into separate modes; further work would integrate the different capabilities to accomplish more complicated tasks. Overall, this work makes progress towards a flexible swarm control that allows user control with minimal difficulty and is robust to the many hardware issues faced in real-world use.
    
\section*{Acknowledgments}
 This material is based on work supported by the Defense Advanced Research Projects Agency (DARPA) project under a DARPA project (funded through Space and Naval Warfare Systems Center Pacific grant N660011924024) and by the National Science Foundation under Grants CNS 1837515. The views, opinions and/or findings expressed are those of the author and should not be interpreted as representing the official views or policies of the Department of Defense or the U.S. Government.

\balance
\bibliographystyle{plainnat}
\bibliography{references}
\end{document}